# Longitudinal Evaluation of Child Face Recognition and the Impact of Underlying Age

Surendra Singh
Clarkson University
New York, USA
sursing@clarkson.edu

Keivan Bahmani
Clarkson University
New York, USA
bahmank@clarkson.edu

Stephanie Schuckers
Clarkson University
New York, USA
sschucke@clarkson.edu

## Abstract

*The need for reliable identification of children in various emerging applications has sparked interest in leveraging child face recognition technology. This study introduces a longitudinal approach to enrollment and verification accuracy for child face recognition, focusing on the YFA (Young Face Aging) database collected by Clarkson University's CITeR research group over an 8-year period, at 6-month intervals. The dataset includes children ranging from 3 to 18 years of age, comprising 330 subjects with an average of 6 data collections per subject. Our research aims to comprehensively evaluate the performance of state-of-the-art face-matching techniques on the YFA database, assessing the feasibility of recognizing children's faces upon initial enrollment and verifying their identity longitudinally at 6-month intervals. We conduct a comprehensive analysis of the system's accuracy considering multiple age groups. We also investigate the temporal degradation of face recognition accuracy over time. Notably, when comparing the initial enrollment image with longitudinal images over an 8-year period, we observe a decrease in accuracy. The average TAR across all age groups is 98.52% with a FAR of 0.1% with a 2-year age verification gap and drops to 95.68 with a 4-year age gap. However, this rate decreases to 87.24% after a time difference of 6 years and further drops to 71.32% with a time difference of 8 years. The highest drop in accuracy was noticed in the age group of (3-5) years old children and the lowest in (5.5-7) years old. By addressing the challenges and opportunities in child face recognition, this research contributes significantly to the advancement of technology for identifying missing or abducted children and other critical applications requiring dependable biometric recognition in children.*

## 1. Introduction

In recent years, there has been a growing demand for reliable identification of children across various applications, including missing children, border security, humanitarian, and health care. This highlights the need to explore the potential of face recognition technology for children. However, traditional face recognition systems have primarily focused on adults, which poses limitations when applied to children due to the unique characteristics of juvenile facial features and how they change over time [16].

Aging in biometric features results in performance degradation for biometric recognition systems [9]. Unlike factors such as lighting or pose that contribute to variability within an identity, aging presents an unavoidable aspect that cannot be controlled during the image capture process [8]. While contemporary Face Recognition (FR) systems, utilizing deep Convolutional Neural Network (CNN)-based approaches, demonstrate robust performance across various poses, illumination, and facial expressions, they still face challenges associated with aging. These systems experience a significant decrease in accuracy exceeding 10% when confronted with substantial age disparities during evaluation [20].

To address this gap, this study is focused on child face recognition, with a specific emphasis on verification accuracy over time using a novel longitudinal dataset. The Young Face Aging (YFA) database contains face images of children captured at 6-month intervals from 2016 to November 2023 for children from 3 to 18 years of age. By leveraging this longitudinal dataset, our research aims to provide a comprehensive evaluation of state-of-the-art face recognition focused on the unique challenges in children.

This longitudinal approach enables us to study the performance impact of the changes in facial appearance that occur as children age. Furthermore, we split the YFA database into different age groups to analyze the pattern of accuracy among different ages. Our objective is to discern whether there exists a consistent trend in accuracy across

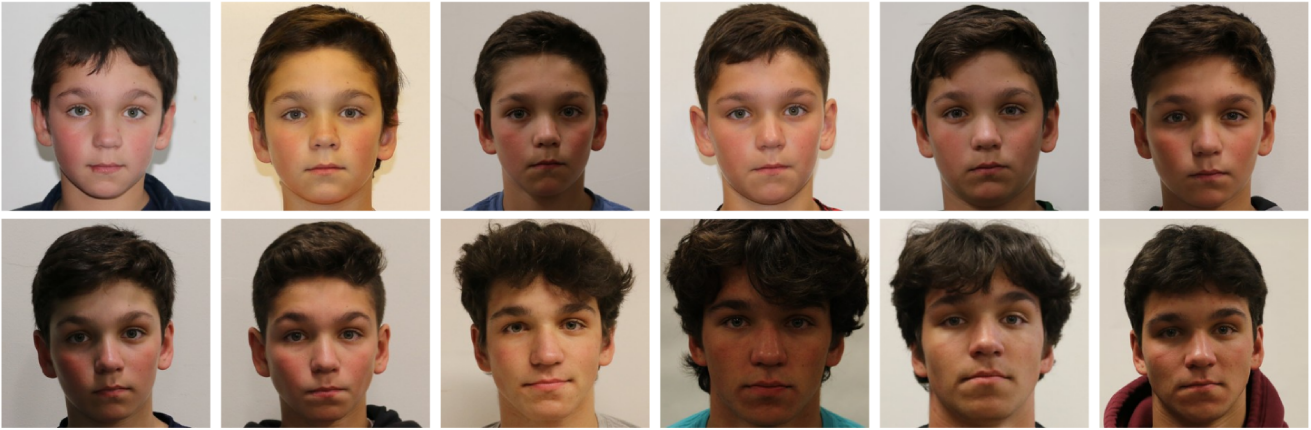
Figure 1. Age progression of a subject in the YFA database from 10 years to 17 years at 6-month intervals.

all age groups, or if certain age groups exhibit a substantial decline in accuracy compared to others. Our contributions are listed as follows.

- We study the performance for increasing the time between as an enrollment and consecutive verification sample over a 6-month time intervals, up to 8 years. Children's ages range from 3 to 18 years, with 330 subjects and an average of 6 collections for each subject. By using this method, our main goal is to gain insights into the performance and reliability of the enrollment and verification system over time, particularly in the context of age progression. Figure 1 shows an example of the image quality of subjects over time.

- Additionally, we have conducted a thorough analysis of the results based on age groups spanning two years e.g (3-5, 5.5-7 years). This detailed examination has enabled us to identify subtle patterns and trends in the system's ability to match identities across different age brackets. By breaking down the data in this manner, we have gained a more comprehensive understanding of the system's effectiveness in accurately verifying identities across different stages of child development.

- We evaluated the True Acceptance Rate (TAR) of the system for each age group using bootstrapping. This process allows us to generate upper and lower bounds on the observed performance and better understand the significance of the difference when changes are observed.

## 2. Related Work

Table 1 provides details of numerous studies in the field of face aging for children that have highlighted the detrimental impact of large age intervals on the accuracy of FR systems. These intervals, are typically measured over many years and are often "in the wild" datasets and/or have a large time gap between collections. There are several constrained datasets where the time gap is more controlled. Most have shorter time period than our dataset ([3], [6], [15]). The most similar dataset is ECLF [5] database which has an average time span of 3.5 years with a 1-year gap between collections with a maximum span of 6 years. The YFA database used in this research has a collection time interval of 6 months over a maximum span of 8 years.

Best-Rowden *et al.* [3], built a Newborn, Infants, and Toddlers Longitudinal (NITL) database of facial images to explore FR in children as they age. According to the study, the accuracy of FR remained high within the same session but decreased significantly across different sessions. This emphasizes the need for further research to improve cross-session recognition accuracy for children. The study provides a comprehensive evaluation of face matching using the NITL database, exploring the possibility of FR for children as they age. The results suggest that current FR technology may not be reliable enough for very young children, but it could be feasible for those enrolled at 3 years of age or older. With a COTS face matcher, the research achieved a TAR of 60.94% at FAR of 0.1% for an age gap between 3-5 years.

Chandaliya *et al.* [5], research compiled a longitudinal database of Indian children (ages 2 to 18, encompassing both boys and girls) using with and without face masks. The findings indicate a significant decline in the performance of facial recognition systems due to aging when masks are utilized. The research utilized the Children Longitudinal Face (ECLF) dataset, comprising 26,258 facial images belonging to 7,473 subjects aged between 2 and 18 years. On average, each subject contributed 3 images, acquired over an average time lapse of 3.35 years. The no-mask dataset achieved an average identification accuracy with a 1-year time interval and a 1-6 year time gap across FaceNet, PFE, ArcFace, and COTS models of 83.79% at 0.01% FAR.

Srinivas *et al.* [16], observed a discernible bias when comparing performance metrics between child and adult face datasets. Their study emphasizes the necessity for a thorough investigation into the implications of FR systems for children. Despite the scarcity of scholarly articles addressing this issue, their contribution to expanding and diversifying the ITWCC dataset suggests a growing interest and potential for further research in this field. The study relied on the ITWCC-D1 database, encompassing 745 subjects with 7,990 images and an age range spanning from 0 to 32 years. They achieved 75.9% TAR at 0.1% FAR and with score level fusion 78.2% TAR at 0.1% FAR. The research was conducted using 8 face-recognition systems. However, the verification accuracy does not account for the age of the subjects at enrollment and verification.

Table 1. Comparison of face recognition datasets utilized by researchers in child face recognition.

| Datasets | Subjects | Samples | Environment | Ages | Time Gap | Collection Period |
|---|---|---|---|---|---|---|
| ITWCC-D1 [16] | 745 | 7,990 | Wild | 0yrs-32yrs | - | - |
| NITL [3] | 314 | 3,144 | Wild | 0-4 years | 6 Months | 1 Years |
| AgeDB [11] | 568 | 16,488 | Wild | 1-101 | Varied | - |
| Morph II [13] | 13,000 | 55,133 | Constrained | 16 -77 Years | 0-5 Years | 36 Years |
| ECLF [5] | 7,473 | 26,258 | Constrained | 2-18 Years | 1 Year | - |
| CLF [6] | 919 | 3,682 | Constrained | 2-18 years | 2-4 Years | 7 Years |
| CMBD [15] | 141 | 2,590 | Constrained | 18months - 4 Years | Months apart | - |
| YFA (Ours) | 330 | 3,831 | Constrained | 3-18 years | 6 Months | 8 Years |

Deb *et al.* [6], present a longitudinal study of FR performance on the Children Longitudinal Face (CLF) database containing 3,682 face images of 919 subjects, in the age group 2-18 years. Each subject has at least four face images acquired over a time span of up to 6-years. In this study, the researchers found out that the accuracy of face recognition decreases over time. Initially, they achieved an accuracy of 83.77% True Acceptance Rate (TAR) and 0.1% False Acceptance Rate (FAR) with a 1-year time-lapse, but this decreased to 59.80% TAR after 3 years. In this research, FaceNet [14] serves as the face matcher, an older face-matching algorithm compared to MagFace [10], which we employ in our study.

Research conducted by Siddiqui *et al.* [15], a novel representation learning algorithm is proposed to extract distinct and invariant features from facial images of newborns and toddlers. This approach is intended to inform the design of an efficient face recognition algorithm tailored specifically for this age group. The CNN architecture proposed in the study achieves a rank-1 identification accuracy of 62.7% for single gallery newborn face recognition and 85.1% for single gallery toddler FR.

A study conducted by Yau *et al.* [18], significant variations in authentication robustness between age groups are demonstrated. The study was conducted using the AgeDB and MorphII databases, encompassing various age groups. Researchers categorized participants into different age groups with a 10-year age gap and then conducted overall accuracy comparisons within each age group.

Bahmani *et al.* [2] conducted a study on children FR, utilizing the YFA database. The study entailed a comparative analysis between YFA and various publicly available cross-age adult datasets to assess the impact of age disparities on both adults and children. The research findings reveal a significant and consistent decline in match scores, with increasing age gaps between gallery and probe images in children, even over short intervals such as 6 months. They use multiple face-matching algorithms Facenet-V1[14], Facenet-V2[14], VGGFace [12], VGGFace2 [4], ArcFace [7], ArcFace-Focal [17] and MagFace [10]. With MagFace they achieved 98.3% and 94.9% TAR at 0.1% FAR over 6 and 36 months age gaps. Considering these results, we used MagFace as the FR model, employing an extended YFA database for this study. Previous research utilized the YFA database with collection periods of up to 3 years, whereas we expanded our database to include collection periods of up to 8 years.

In our research, the YFA dataset enables consideration of the accuracy for more controlled time intervals, as well as high-quality images. YFA covers ages 3-18, as well as increases in time between enrollment and verification for 6-month increments up to 8 years. Through this work, we aim to provide a more comprehensive understanding of how the performance of the system evolves with the aging process, offering valuable insights into the effectiveness and reliability of FR technology across various age demographics.

## 3. Methodology

The primary objective of this research is to develop and evaluate a longitudinal age enrollment and verification system using the YFA dataset. The methodology employed involves two main stages: enrollment and verification. *Enrollment Stage:* Any image in the dataset may be considered an enrollment stage, where subjects are categorized into specific age brackets. *Verification Stage:* Following the enrollment stage, all subsequent collections of the YFA dataset are utilized for verification purposes. These collections are considered as verification samples, with the time interval between collections increasing as the subject's age progresses. This approach allows for the evaluation of the system's performance over time, considering the aging effects on facial appearance.

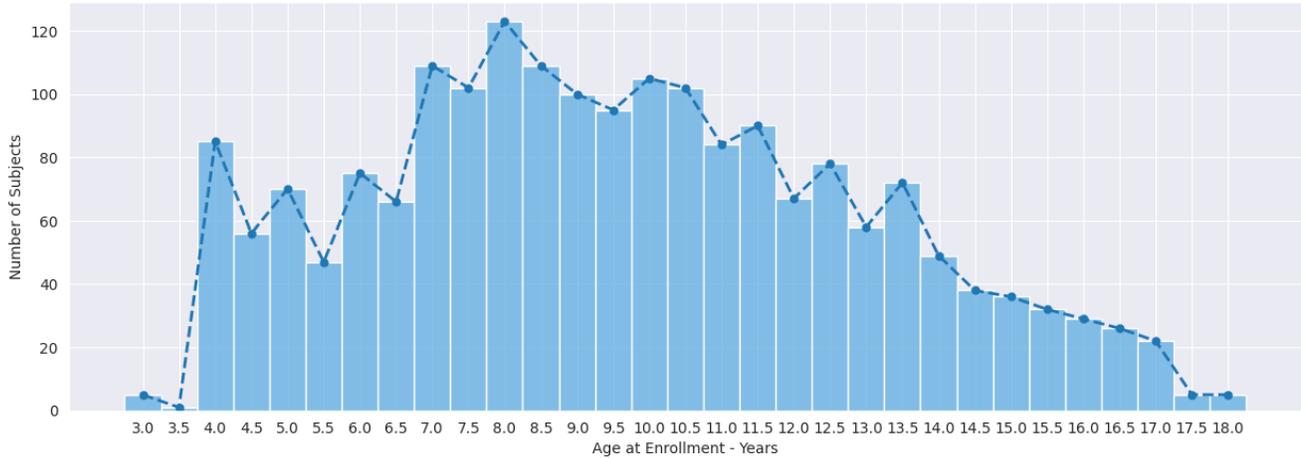

Figure 2. Number of subjects at each age in the YFA Database. The same subject might have been captured at different ages.

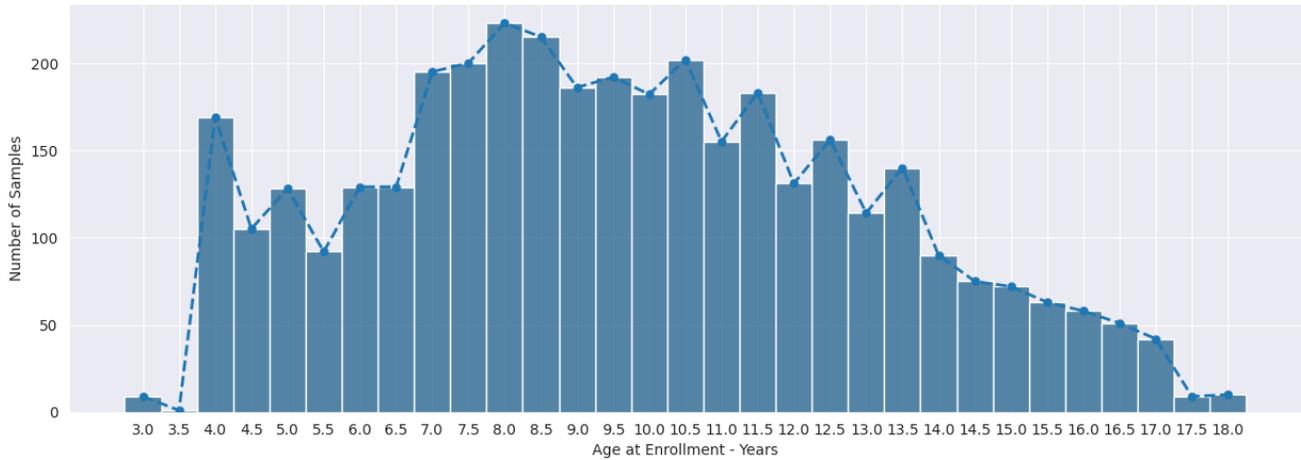

Figure 3. Number of images at each age in the YFA Database. Typically each subject provided 2 images at each session, but there may be more or less in a few cases due to quality issues or extra samples taken.

### 3.1. Database Overview

The YFA database serves as the foundation for this research. The research team collaborates with the local elementary, middle, and high schools to identify and enroll subjects for voluntary participation, following an approved IRB protocol. Collections occur at 6-month intervals. Each year, new subjects are added at ages 3-5 (i.e., pre-K and Kindergarten students. Each collection contains facial images of subjects, along with their corresponding age. Figures 1 illustrate the age progression of individual subjects within the YFA database, showcasing the development from 10-17 years of age, with intervals of 6 months. The database contains 3831 samples from 330 subjects collected in a controlled environment with a time-lapse of 6 months over 8 years. Figure 2 depicts the statistics of the YFA. Samples are captured from 3-18 year old children. Images are captured using a DSLR camera with a resolution of 3648 by 5472 pixels. The image acquisition is conducted under consistent indoor lighting conditions, encouraging neutral expressions, and minimizing variations in the subject's pose. Manual annotation by human annotators was performed and extremely blurry images and challenging poses were excluded from the database. Each subject is captured at least twice during each session. Figure 3 illustrates the number of samples available at each age. This dataset exhibits the highest number of collections for subjects aged 8 years and the lowest number of samples for subjects aged 3.5 years. Ages are approximate based on the school grade of enrollment as some subjects declined to provide birthdate.

Table 2. Overall performance (TAR% at 0.1% and 0.01% FAR) on YFA database with cross-age matching.

| Model | TAR @0.1% FAR | Threshold | TAR @0.01% FAR | Threshold |
|---|---|---|---|---|
| MagFace | 95.48 | 0.45 | 82.25 | 0.56 |

## 4. Face Detection and Recognition Models

MTCNN (Multi-Task Cascaded Convolutional Neural Network) face detection model, as referenced in [19] was employed to accurately detect and align faces. Additionally, we resize each cropped face to meet the requirements of the facial recognition matcher's input specifications at $224 \times 224$ pixels. This establishes a consistent foundation for the comprehensive analysis and evaluation of our research.

The performance of the longitudinal age enrollment and verification system is assessed based on face-matching accuracy across different time intervals. Bahmani *et al.* [2], introduce the YFA dataset for analyzing the performance of FR systems over short age gaps in children. They use multiple face-matching algorithms; Facenet-V1[14], Facenet-V2[14], VGGFace [12], VGGFace2 [4], ArcFace [7], ArcFace-Focal [17] and MagFace [10]. The best performance was with MagFace which achieved 98.3% and 94.9% TAR at 0.1% FAR over 6 and 36 months age gaps. Based on these results, we used MagFace as a FR model in this study.

## 5. Experimental Setup

To facilitate analysis of the longitudinal aging factor, we categorize the longitudinal age of subjects as enrollment and verification. Figure 2 details the available subjects of the YFA dataset for experiments. Upon initial enrollment of a subject into the system, we determine their age based on the enrollment date, given that the YFA database records data at 6-month intervals, we increment the subject's age by 6 months with each subsequent data collection. Notably, the youngest age recorded in the database is 3 years. Consequently, we designate all subjects at the age of 3 years as enrollment samples. Following a period of 6 months, when the subject's age is 3.5 years, we classify them as verification samples for further analysis. This repeats for all images from that subject. In this systematic approach, each image can be either an enrollment or verification image (with the exception of the first image captured for a child can only be an enrollment image). Figure 3 provides the number of samples available in each enrollment period. The largest number of subjects in the database are 8 years old and the lowest number of subjects is 3.5 years old.

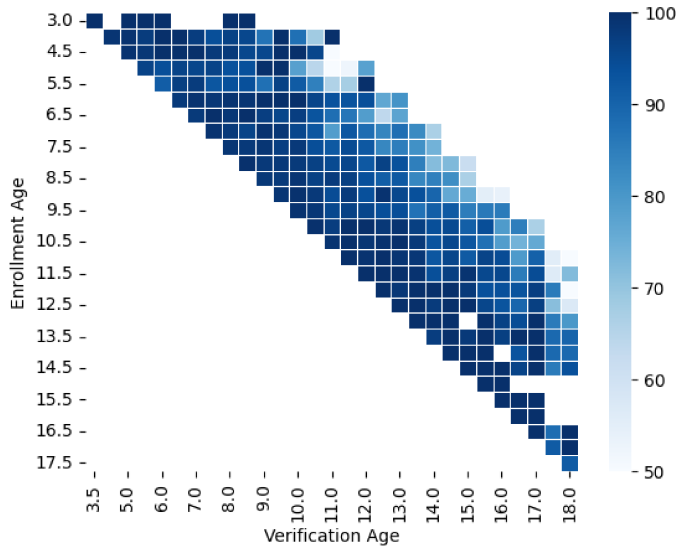

Figure 4. Heatmap of TAR at 0.1% FAR for increasing time periods between enrollment and Verification for each age. The heatmap illustrates a notable decline in TAR as the time between enrollment and verification increases.

## 6. Results & Discussion

### 6.1. Enrollment and Verification Scenarios

We explore overall cross-age performance in the range of up to 8 years within the YFA database. Table 2 provides a detailed depiction of the TAR computed for the MagFace [10] face matcher model across two FAR thresholds, specifically at 0.1% and 0.01%. Specifically, we set a fixed threshold value of 0.45 at a FAR of 0.1% and 82.25 %TAR at 0.01% FAR with 0.56 threshold. In our further analysis, we keep the 0.45 threshold at 0.1% FAR to calculate the TAR performance of each subset of age-wise enrollment and verification. Figure 3 provides the statistics of available samples in each subset of ages. To evaluate the longitudinal performance we compare the age of the subject during the first enrollment to the age of the subject for each subsequent verification. The YFA database collection period is 8 years so the maximum difference between enrollment and verification is 8 years (*i.e* This longitudinal approach allows for a nuanced understanding of how the recognition system performs over time, particularly in relation to the aging process of individuals.

By tracking the subject's ages over the 8-year collection period, we were able to observe trends in the system's performance, specifically focusing on the TAR across different age groups and over increasing time between enrollment and verification. Figure 4 visually represents our findings. Our analysis reveals a decline in TAR, particularly beyond the fourth year after enrollment across all age groups. For certain age comparisons, e.g. age 13 compared to age 16, there are no images that meet that criteria and thus it is represented as a white box. There is a noticeable decrease in TAR as age increases, particularly when it has been more than 4 years since enrollment.

Table 3. TAR% at 0.1% FAR for each enrolment age (rows) for increasing time gaps between enrollment and verification (columns). Each verification column consists of TAR% upper and lower bound calculated based on bootstraping and column N represents available subjects for that time gap.

| | TAR% at 0.1% FAR for different time gaps between enrollment and verification | | | | | | |
|---|---|---|---|---|---|---|---|
| Enrollment Age | 0.5-2 Years | N | 2.5-4 Years | N | 4.5-6 Years | N | 6.5-8 Years | N |
| 3-5 Years | 97.8 [96.8, 98.7] | 88 | 96.3 [95.1, 97.5] | 65 | 86.7 [84.4, 88.8] | 55 | 63.1 [60.2, 66.1] | 20 |
| 5.5-7 Years | 97.9 [97.0, 98.7] | 133 | 95.2 [93.8, 96.5] | 82 | 85.9 [83.6, 87.9] | 63 | 80.2 [77.8, 82.5] | 43 |
| 7.5-9 Years | 98.3 [97.6, 99.0] | 161 | 95.7 [94.4, 96.9] | 99 | 85.7 [83.6, 88.1] | 88 | 65.6 [62.5, 68.4] | 51 |
| 9.5-11 Years | 98.9 [98.3, 99.5] | 141 | 95.4 [94.1, 96.6] | 112 | 88.7 [86.4, 90.5] | 89 | 75.5 [72.9, 77.9] | 39 |
| 11.5-13 Years | 99.7 [99.4, 100.0] | 114 | 95.8 [94.6, 97.0] | 65 | 89.2 [87.2, 91.0] | 41 | 72.2 [69.4, 75.1] | 4 |

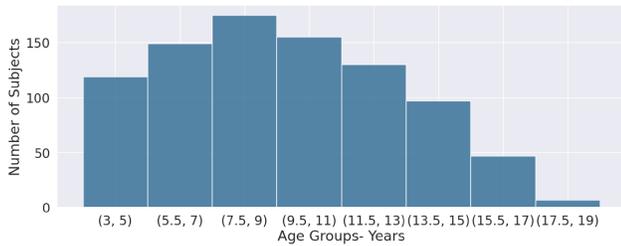

Figure 5. Distribution of subjects for age groups where each age group spans 2 years

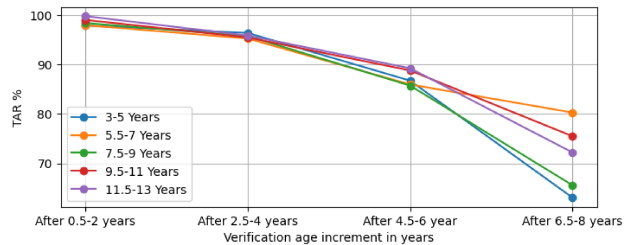

Figure 6. TAR% for each enrolment age (lines) for increasing time between enrollment and verification (x-axis).

This decline in TAR could stem from various factors, including changes in facial features due to natural aging processes. However, poor performance could also be due to other factors, such as pose, illumination, expression, or environmental influences. We controlled for these factors as best as possible, but there are likely still influences. Understanding these trends is crucial for developing strategies to improve the longevity and effectiveness of biometric FR systems.

## 6.2. Analysis for age groups

Additional investigation was performed in the observed trend of declining TAR to see if the underlying age of enrollment has an impact. In other words, we performed further analysis to determine if there is a difference between a subject that enrolls at age 5 and verifies at age 10 versus a subject that enrolls at age 10 and verifies at age 15, for example. We created subsets of the database, with subjects categorized into groups based on a 2-year age interval. This subset approach allowed for a examination of potential age-related trends in the performance.

The distribution of subjects for each of the age groups is depicted in Figure 5. For the analysis, each group is matched over multiple 2-year time intervals. For instance, subjects enrolled between the ages of 3-5 years were compared against every image collected from those subjects between 6-month to 2 years after enrollment and similarly from (2.5-4) year time difference up to 8 years of time gap. This methodology enabled tracking of performance trends across various age groups at increasing time of verification after enrollment

Table 3 further highlights the decline in the TAR as time after enrollment increases. For less than 2 years from enrollment, Age groups (3-5, 5.5-7, 7.5-9, 9.5-11) had a TAR of 97.8 - 98.9% and a TAR of 99.75 for age group (11.5-13). After 2.5 to 4 years, age groups (3-5, 5.5-7, 7.5-9, 9.5-11) observed nearly 1-3% drop in TAR and (11.5-13) close to 5% drop in TAR. After 4.5 to 6 years, all age groups observed 10% drop in TAR. After 6 years from enrollment, there was a subsequent sharp decline. Figure 6 visually represents the trend on the drop of TAR across different age groups. This observed pattern persists consistently across nearly all age groups. Through systematic TAR evaluation within each age group, we can analyze whether the observed TAR decline is uniform across all age groups or if certain age groups experience more pronounced performance deterioration over time. We noticed that the enrollment age groups (3-5) and (7.5-9) years have a larger drop in TAR compared to other age groups. For the enrollment age group (3-5) with a verification age gap of (0.5-2) years, the TAR decreased from 97.8% to 63.1% for a verification age gap of (6.5-8) years. Similarly, for the enrollment age group (7.5-9) with a verification age gap of (0.5-2) years, the TAR decreased from 98.3% to 65.6% for a verification age gap of (6.5-8) years. With age group (5.5-7) years, the TAR decreased from 97.9% to 80.2% for a verification age gap of (6.5-8) years. Noticed that age groups (3-5) and (7.5-9) years had a 20-23% drop in TAR after 6 years of a verifi-cation age gap whereas age group (5.5-7) had a 5% drop in TAR.

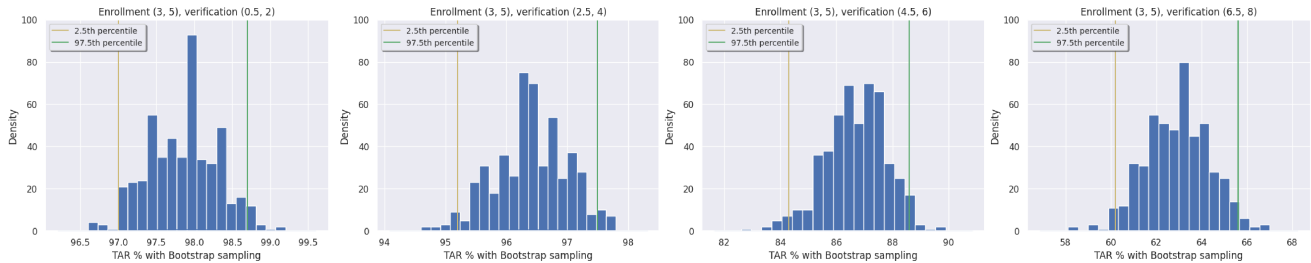

Figure 7. Bootstrap visualization of a two-sided 95% confidence interval for TAR, bounded by the 2.5th and 97.5th percentiles. Fewer subjects results in wider confidence intervals, particularly for longer time intervals (6.5 to 8 years from enrollment).

## 6.3. Confidence intervals based on bootstrapping for performance evaluation

Based on the findings in the previous section, we noticed that subjects are not evenly distributed across the various age groups and time gaps. Confidence intervals were constructed using bootstrap resampling process to quantify the uncertainty surrounding key statistical estimates [1]. Bootstrapping is a statistical resampling technique used to estimate the sampling distribution of a statistic by repeatedly sampling with replacement from the observed data. It allows for the assessment of variability and uncertainty in a sample estimate without assuming a specific parametric distribution. Figure 7 is a visual representation of the resampling distribution of data across different age groups. For instance, a two-sided 95% confidence interval was computed using the bootstrap percentile method and represents the TAR range in which the true parameter value is expected to lie with 95% probability. The lower bound of the confidence interval is the TAR value at the 2.5th percentile. Conversely, the upper bound is the TAR value at the 97.5th percentile.

Table 3 provides lower and upper bounds for each TAR. This confidence interval estimation accounts for the variability inherent in the data and allows comparison of the trends to support statistical inference and decision-making. For instance, the enrollment age (3-5) with a verification age of (6.5-8) years consists of only 20 subjects. The 95% confidence interval for a TAR range is from 60.2 to 66.1. This analysis provided a better understanding of the plausible range of TAR values. With bootstrap resampling with various age groups, particularly focusing on the verification age gap of (6.5-8) years, we observed that the enrollment group aged (5.5-7) years did not experience as significant of a decline in TAR compared to other groups after 6 years of verification gap, TAR decreased from 85.9% to 80.2% until 8 years verification age gap. However, we found that the enrollment age groups of (3-5) years and (7.5-9) years showed lower TAR, particularly compared to the verification group of (4.5-6) years to (6.5-8) with 63.1% and 65.6% TAR respectively.

Comparison of performance is provided in Table 4. Most prior work has poor performance either because the dataset are in the wild and have poor-quality images or the algorithms are older. In a recent study with a 6-year time difference [5], ArcFace and a COTS model achieved a TAR of 98.32% and 98.99% respectively, with a 0.1% FAR. This performance is better achieved in the YFA database even though MagFace has been shown to outperform ArcFace in our dataset. The main difference between the datasets are race and ethnicity. The database used [5] primarily consists of Indian children, in contrast to the YFA database, which contains a high proportion of children which are white. There is not further analysis of the age or age group of the subject during enrollment and further verification matching. Thus the difference could also be related to the distribution of ages.

## 7. Conclusion and Future Work

In conclusion, the study's findings underscore the importance of longitudinal assessments in evaluating the performance of face recognition systems, particularly in understanding how age-related factors impact their accuracy and reliability over time. Further research and refinement of algorithms may be necessary to address the observed decline in TAR and enhance the robustness of FR technology across diverse age demographics.

We performed an evaluation of the MagFace facial recognition algorithm for the YFA database, encompassing subjects aged 3 to 18 years, and up to 8 years between enrollment and verification samples. Our comprehensive analysis includes a breakdown of results based on age groups spanning two years. The average TAR in the validation group aged (0.5-2) years was 98.52%, whereas it decreased to 95.68% in the validation group aged (2.5-4) years. We find a drop in accuracy after a 4-year age difference among subjects and a sharp decline after a 6-year age difference with the average drop in TAR from 87.24% to 71.32%. Our analysis, focusing on specific age groups, reveals a notable trend indicating a decrease in TAR over time following enrollment. We observed the most substantial decline in accuracy among children aged 3-5 years old, where the TAR dropped to 63.1%. This decline occurred particularly within a verification age gap of (6.5-8) years. Conversely, the age group of (5.5-7) years old exhibited the highest TAR 80.2%. Our examination further reveals that the accuracy performance does not exhibit consistent trends with increasing age within enrollment age groups. For instance, within the verification age group of (6.5-8) years for children aged (3-5) years, the TAR dropped to 63.1%. However, for the age group of (5.5-7) years, the accuracy notably increases to 80.2%. This trend reverses once more for the (7.5-9) age group, with accuracy dropping to 65.6%. Interestingly, there is a slight uptick in accuracy observed in the subsequent age groups. These fluctuations underscore the complexity of age-related dynamics in biometric verification accuracy and suggest the need for tailored approaches to address variations across different age ranges. These findings not only contribute to a deeper understanding of facial recognition technology's for children and can inform its implementation. Moving forward, our research sets a foundation for continued exploration and refinement of FR systems to ensure their efficacy and fairness across all age demographics.

Table 4. Comparative analysis of prior child face recognition studies: database, model, and accuracy.

| Database | Longest time gap | Time interval | Accuracy | Model |
|---|---|---|---|---|
| ECLF [5] | 6 years | 1 year | TAR at 0.1% FAR | FaceNet: 84.55 |
| | | | | PFE: 98.90 |
| | | | | ArcFace: 99.38 |
| | | | | COTS: **99.62** |
| ITWCC-D1 [16] | - | - | TAR at 0.1% FAR | FR Model: COTS |
| | | | | FR-A: 0.676 |
| | | | | FR-B: 0.598 |
| | | | | FR-C: 0.463 |
| | | | | FR-D: 0.434 |
| | | | | FR-E: **0.759** |
| | | | | FR-F: 0.738 |
| | | | | FR-G: 0.718 |
| | | | | FR-H: 0.695 |
| NITL [3] | 2 years | 1 year | TAR at 0.1% FAR | COTS: **60.94** |
| CLF [6] | 3 years | 3 year | TAR at 0.1% FAR | COTS: 49.33 |
| | | | | FaceNet: 59.80 |
| CMBD [15] | - | - | Rank-1 Accuracy | PCA: 38.8 |
| | | | | LBP : 28.8 |
| | | | | LDA : 71.3 |
| | | | | Fine-tuned VGG-Face: 83.0 |
| | | | | Triplet CNN : 72.7 |
| | | | | Proposed CNN: **85.1** |
| YFA [2] | 3 years | 6 months | TAR at 0.1% FAR | Facenet-V1: 76.0 |
| | | | | ArcFace : 81.1 |
| | | | | ArcFace-Focal : 91.6 |
| | | | | MagFace: **94.9** |
| Ours -YFA extended | 8 years | 6 months | TAR at 0.1% FAR | MagFace: **95.48** |

The research faces several limitations concerning its database. Although the study encompasses children aged 3 to 18 years, the distribution of subjects within this age range is uneven, potentially impacting the generalizability of findings. While the collectors attempted to control various quality factors, variations in facial movement, angle, and facial expressions were present. All collections were performed in a classroom environment with the lights on. However, there are still inconsistencies in lighting across collections. Moreover, disparities in subjects' appearance, such as some wearing glasses and caps, further complicate the recognition process. The lack of demographic diversity within the database, particularly in terms of ethnicity, poses a significant limitation to the generalizability of the findings. In this study, we use the MagFace face-matching algorithm. Other face matchers or fine-tuning the existing algorithm can impact the results. The study aims to assess children's facial recognition for critical applications, such as identifying missing or abducted children. However, the findings are in a controlled environment and performance may be lower in real-world scenarios due to factors like image quality variations, environmental conditions, and operational constraints.